# Improving the performance of optical inverse design of multilayer thin films using CNN-LSTM tandem neural networks

Uijun Jung[1], Deokho Jang[1], Sungchul Kim[2], and Jungho Kim[1, *]

[1]Department of Information Display, Kyung Hee University, Seoul, 02447, Republic of Korea
[2]Department of Information and Communications Engineering, Myongji University, Yongin-si, 17058, Republic of Korea

*junghokim@khu.ac.kr

## ABSTRACT

Optical properties of thin film are greatly influenced by the thickness of each layer. Accurately predicting these thicknesses and their corresponding optical properties is important in the optical inverse design of thin films. However, traditional inverse design methods usually demand extensive numerical simulations and optimization procedures, which are time-consuming. In this paper, we utilize deep learning for the inverse design of the transmission spectra of $SiO_2$/$TiO_2$ multilayer thin films. We implement a tandem neural network (TNN), which can solve the one-to-many mapping problem that greatly degrades the performance of deep-learning-based inverse designs. In general, the TNN has been implemented by a back-to-back connection of an inverse neural network and a pre-trained forward neural network, both of which have been implemented based on multi-layer perceptron (MLP) algorithms. In this paper, we propose to use not only MLP, but also convolutional neural network (CNN) or long short-term memory (LSTM) algorithms in the configuration of the TNN. We show that an LSTM-LSTM-based TNN yields the highest accuracy but takes the longest training time among nine configurations of TNNs. We also find that a CNN-LSTM-based TNN will be an optimal solution in terms of accuracy and speed because it could integrate the strengths of the CNN and LSTM algorithms.

## 1. Introduction

Thin films are widely used in various applications such as light-emitting diodes, solar cells, and flat panel displays [1-6]. When light passes through a thin film, the transmission/reflection spectrum greatly depends on the refractive index, extinction coefficient, and thickness of the material that makes up the film [3]. The corresponding transmission/reflection spectrum can be calculated through the transfer matrix method (TMM) [3]. The more complex the structure of a thin film, the more time-consuming and expensive it is to design such that the optimization of thin-film optical design based on deep learning has been actively pursued recently [7,8].

Deep learning is a branch of machine learning that uses artificial neural networks to learn from data [9,10]. When deep learning is applied to the optical design of thin films or nano photonics, two types of neural networks have been implemented. The first one is the forward neural network (FNN), which takes the optical structure such as the layer thickness as inputs and predicts its optical properties such as transmission spectrum as outputs. The other corresponds to the inverse neural networks (INN), which predict the optical structure from its optical properties [11-14]. The INN suffers from slow convergence and low accuracy due to a so-called one-to-many mapping problem, which occurs when different structures of the multilayer thin-film have nearly the same optical properties [15]. The tandem neural networks (TNN), which is implemented through a back-to-back connection of an INN and a pre-trained FNN, have been proposed to alleviate the aforementioned one-to-many mapping problem [15,16]. Because TNN is trained to minimize the error between the input target spectra and output predicted spectra, the TNN can avoid the one-to-many mapping problem and can provide accurate inverse design results [15].

In general, TNN has been implemented based on a back-to-back connection of two multi-layer perceptron (MLP), where the first and second MLPs work as the INN and the FNN, respectively [15-20]. This conventional MLP-MLP-based TNN is effective when the number of thin-film layers for inverse design is small, but its performance dramatically drops as the number of layers increases. To improve the performance of the deep-learning-based inverse design, other neural network algorithms such as convolutional neural networks (CNNs) and long short-term memory (LSTM) were used [21-26]. However, there has been no comprehensive performance comparison of hybrid network structures of TNNs such as a MLP-CNN-based, a CNN-LSTM-based, or an LSTM-LSTM-based TNN. In this paper, we implement nine network structures of TNNs and compare their accuracy and training times for the inverse design of multilayer thin films. We find that LSTM-LSTM-based TNN gives the highest accuracy with the longest training time. In addition, we show that a CNN-LSTM-based TNN gives the second highest accuracy while reducing the training time by a factor of four compared to the LSTM-LSTM-based TNN. The CNN-LSTM-based TNN can provide an optimal solution in terms of accuracy and speed because it gives the better accuracy than the conventional MLP-MLP-based TNN with acceptable training time.

## 2. Generation of the training datasets and deep learning algorithms of MLP, CNN, and LSTM

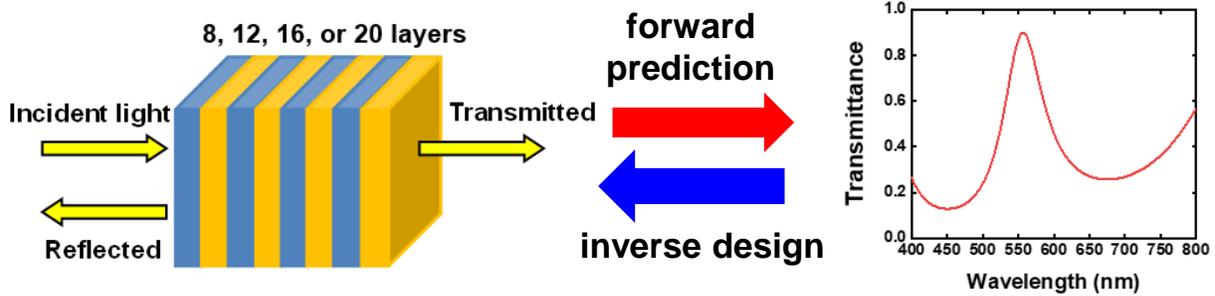

**Fig. 1.** Schematic diagram of the multilayer structure and corresponding transmission spectrum of $SiO_2/TiO_2$ thin films, which are used as input and output for forward prediction of the transmission spectrum from a sequence of the layer thicknesses. In the inverse design, a sequence of the layer thickness (output) is predicted from the target transmission spectrum (input).

Figure 1 shows a schematic diagram of the multilayer structure and corresponding transmission spectrum of $SiO_2/TiO_2$ thin films, which are used as input and output for forward prediction of the transmission spectrum from a sequence of the layer thicknesses. In the inverse design, a sequence of the layer thickness (output) is predicted from the target transmission spectrum (input). The multilayer thin film is composed of alternating layers of $SiO_2/TiO_2$, and their respective refractive indices are obtained from the literature [27]. For reference, the refractive index values at the wavelength of 600 nm, which is the midpoint of the spectral range of the transmission spectrum, are 1.458 for $SiO_2$ and 2.605 for $TiO_2$, respectively. The multilayer thin film consists of 8, 12, 16, or 20 layers and each layer thickness is randomly chosen within the thickness range of 30-70 nm with the thickness interval of 1.0 nm. For normal incidence of light, the corresponding transmission spectrum ranging from 400 to 800 nm with the wavelength interval of 1 nm is calculated based on the TMM, which is implemented by a self-developed code written in MATLAB. The details about the TMM model and refractive index spectra of the materials are presented in Section 1 of Supplementary Information. Using a personal computer with an Intel Core i5-13600K CPU, NVIDIA RTX 4070Ti GPU, and 64 GB RAM, we perform numerical simulations to generate datasets that include a sequence of the multilayer thicknesses and corresponding spectral transmittance. After the numerical calculations, the multilayer thicknesses are normalized between 0 and 1 using the minmax scaler. It takes approximately 6 hours 20 minutes to generate 100,000 datasets when the number of the $SiO_2/TiO_2$ multilayer is 20.

Figure 2 represents a schematic diagram of MLP, which is an artificial neural network with multiple hidden layers between the input and output layers [10]. The greater the number of hidden layers, the more effectively the model can learn complex patterns in the input data. Input data enters at once and passes through multiple hidden layers, and the final output is calculated based on the weight and bias parameters of hidden layers. This corresponds to forward propagation, during which the model's prediction accuracy is evaluated using a loss function. Subsequently, based on the loss function, the error is propagated backward from the output layer to input layer, which is a process known as backpropagation. During this process, gradients for each weight and bias are calculated, and these are used to update the weights and biases in the direction that minimizes the error. By repeating forward and backward propagation, the model gradually improves its predictive accuracy through learning.

Figure 3 shows a schematic diagram of CNN, which is mainly used in the deep learning application to image or video data. Typical image data is represented in the form of height × width × number of channels, but one-dimensional data, such as transmission spectra, is expressed as the form of length × 1 × number of channels. CNNs use a preprocessing method of convolution, which makes CNNs learn the association between the adjacent pixel using the kernel. Each kernel scans the data to generate a feature map. During this process, the stride determines the movement interval of the kernel, and padding can be applied by adding zeros to the edges to preserve the input size. Then, through the pooling layer, the size of the data is reduced to decrease computational load and prevent overfitting. Max pooling selects the maximum value within the kernel's range, while average pooling selects the average value. The features condensed through convolutional and pooling layers are flattened into a 1D vector and passed to the fully connected layer to calculate the final output. Similar to MLPs, forward propagation and backpropagation are repeated to improve the model's performance.

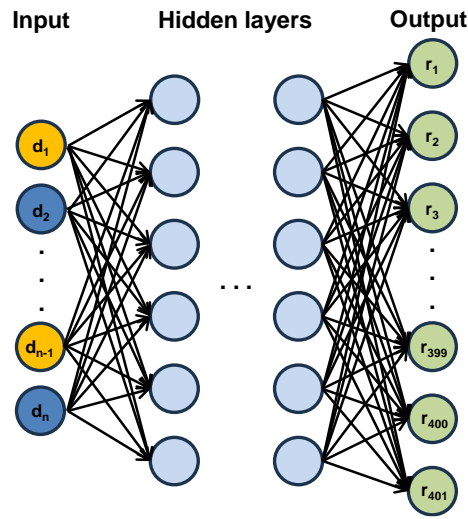

**Fig. 2.** Schematic diagram of MLP.

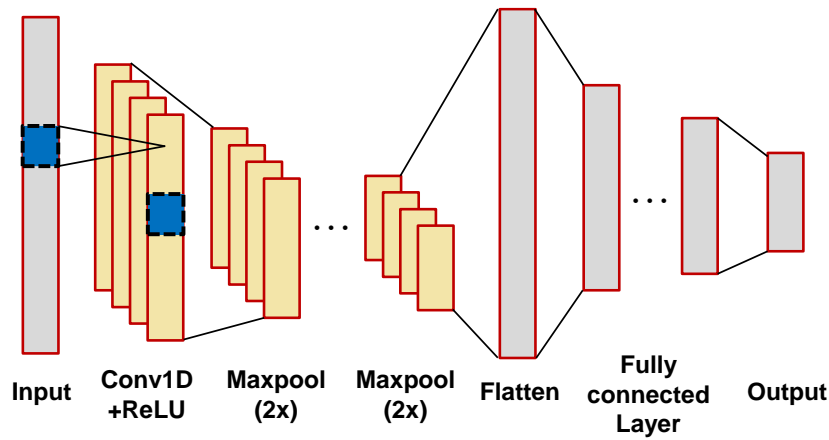

**Fig. 3.** Schematic diagram of CNN.

Figure 4 shows a schematic diagram of LSTM. Recurrent neural networks (RNNs), unlike MLPs and CNNs where the input data enters the hidden layer all at once, are able to make accurate predictions by feeding data sequentially one by one to properly combine new information with existing information. However, when the input data is long, it is difficult to remember the entire information due to gradient decay, and LSTMs were designed to solve this problem [22]. LSTMs are characterized by having memory cells that can retain previous information for a long time, which allows them to learn the associations between the entire input data. An LSTM consists of six parameters and four gates. The cell state is what drives the entire chain, and information is added or deleted by the gates. In this case, the information does not change at all, and the gradient propagates well even if the state has long elapsed. First, forget gate decides whether or not to discard past information, and uses a sigmoid as its activation function, keeping everything with a value of 1 and discarding everything with a value of 0. Second, input gate is for remembering the current information and determines how much to add to the current cell state value. In this case, we mainly use hyperbolic tangent function as an activation function. Next, the values stored in the forget gate and input gate are calculated and updated to the cell state, and then the output gate determines the final output value by deciding how much to subtract from the final value obtained. This value is then passed to the next cell state, and the process is repeated [31,32].

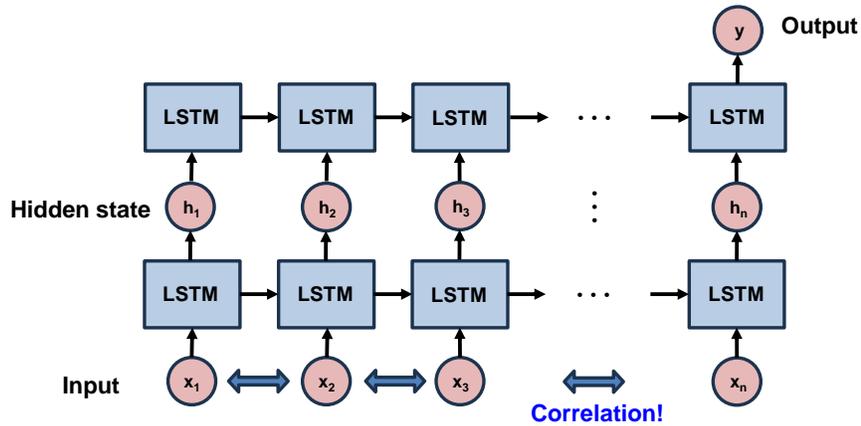

**Fig. 4.** Schematic diagram of LSTM.

## 3. Network structure and training procedure of hybrid TNNs

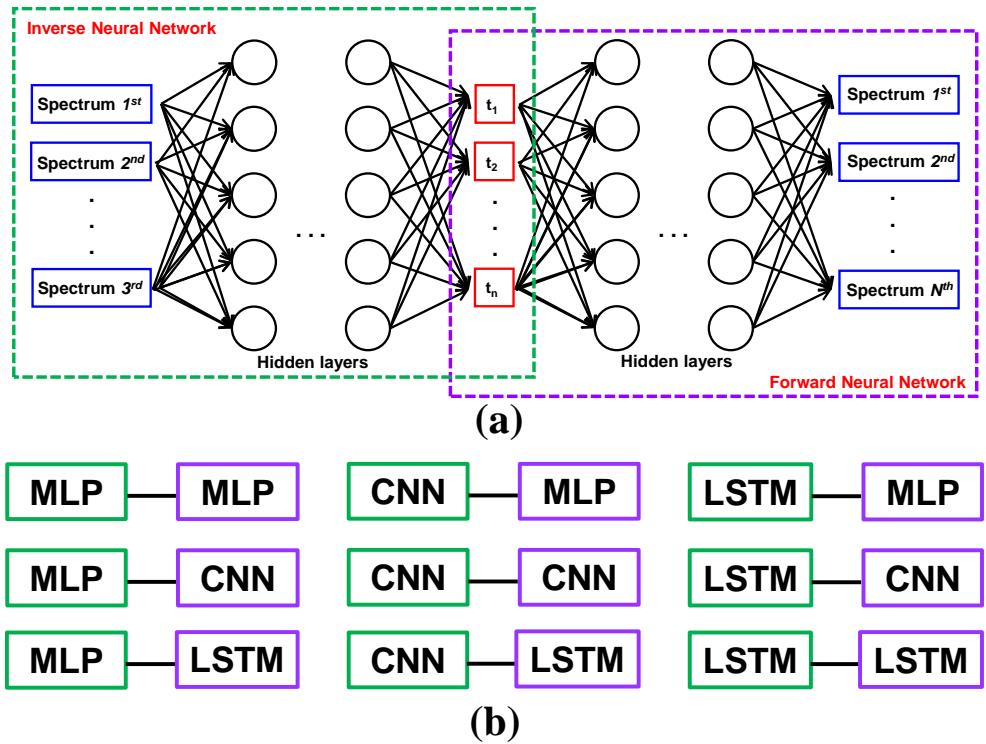

**Fig. 5.** (a) Network structure of the TNN. The purple dashed box indicates the pre-trained FNN, and the green dashed box represents the INN. The hyperparameters of the pre-trained FNN are fixed but those of the INN are optimized to minimize the loss between the target (input) and predicted (output) transmission spectra. The output obtained by the intermediate layer are the inversely-designed multilayer thicknesses for the target transmission spectrum. (b) Schematic diagram of hybrid TNNs having nine different configurations. Three algorithms of MLP, CNN, and LSTM are used as either the INN or the pre-trained FNN to implement the TNN.

Figure 5(a) shows a network structure of the TNN, where the purple dashed box indicates the pre-trained FNN and the green dashed box represents the INN, respectively. In Figure 5(a), the output obtained by the intermediate layer are the inversely-designed multilayer thicknesses for the target transmission spectrum. Figure 5(b) represents a schematic diagram of hybrid TNNs with nine different configurations, where three algorithms of MLP, CNN, and LSTM are used as either the INN or the pre-trained FNN. The hyperparameters of the FNN is optimized to minimize the learning loss between the normalized multilayer thicknesses (input) and transmission spectrum (output) when either MLP, CNN, or LSTM is used as a FNN. After the hyperparameters of the pre-trained FNN are fixed in the TNN, the hyperparameters of the INN inside the TNN, which can also be either MLP, CNN, or LSTM, are adjusted to minimize the learning loss between the target (input) and the predicted (output) transmission spectra. The hyperparameters for each algorithm are adjusted in reference to the literature on inverse design [15,22,23,25]. The same datasets described above are used to train both the FNN and the TNN, separately. Among 100,000 datasets, we use 60% of them for training, 20% for validation, and the remaining 20% for testing.

## 4. Result and Discussion

### 4.1. Forward neural network

**Table 1**
Performance comparison of three FNNs applied to a 20-layer thin film depending on the network's algorithm.

| Algorithm | R2 score | MSE | Training time (s) |
|---|---|---|---|
| MLP | 0.9521 | 0.0010 | 6,818 |
| CNN | 0.9450 | 0.0013 | 16,826 |
| LSTM | 0.9773 | 3.4227e-4 | 23,366 |

Table 1 shows performance comparison of three FNNs depending on the neural network's algorithm of MLP, CNN, and LSTM when the number of the thin-film layer is 20. Each FNN is separately trained to minimize the learning loss that is calculated based on the mean squared error (MSE). In Sections 2 of Supplementary Information, we present more details about the network structure and hyperparameters of three FNNs such as the number of hidden layers, the number of neurons in each hidden layer, and a type of the activation function. We compare the accuracy of the prediction results among the MLP-. CNN-, and LSTM-based FNNs by calculating R2 scores and MSE between the predicted and calculated transmission spectra in 20,000 test datasets. We employed the same calculations three times for each of the three FNN algorithms and summarized the averaged values in Table 1. According to Table 1, the LSTM-based FNN shows the best prediction accuracy due to the inherent memory characteristics of the LSTM algorithm, which enable sequential data processing to effectively integrate past and present training datasets and result in more accurate forward predictions [23,24]. On the other hand, the LSTM-based FNN has the longest training time by a factor of 3.4 compared to the MLP-based FNN. In addition, it is noticeable that the CNN-based FNN has the worst accuracy, which is inconsistent with the well-known fact that CNN generally gives the better accuracy than MLP [22,25]. This discrepancy can be explained as follows. In the application of forward prediction, the input data, which corresponds to an 1 × 20 vector, consists of an independent layer thickness, which is randomly selected within the thickness range. Because there is no correlation between adjacent input values of the FNN, the working principle of the CNN, which makes use of the local correlation between neurons of adjacent layers, will not be effective in the implementation of the CNN-based FNN. Hence, the accuracy of the CNN-based FNN can be worse than that of the MLP-based FNN.

### 4.2. Hybrid tandem neural network

**Table 2**
Performance comparison of nine hybrid TNNs applied to a 20-layer thin film depending on the network's algorithm.

| Algorithm (INN-FNN) | R2 score | MSE | Training time (s) |
|---|---|---|---|
| MLP-MLP | 0.9369 | 0.0013 | 18,780 |
| MLP-CNN | 0.9349 | 0.0014 | 33,014 |
| CNN-MLP | 0.9476 | 0.0010 | 28,204 |
| CNN-CNN | 0.9468 | 0.0010 | 43,311 |
| MLP-LSTM | 0.9512 | 9.4493e-4 | 79,847 |
| CNN-LSTM | 0.9657 | 6.1684e-4 | 67,362 |
| LSTM-MLP | 0.9324 | 0.0013 | 222,933 |
| LSTM-CNN | 0.8135 | 0.0055 | 240,704 |
| LSTM-LSTM | 0.9776 | 3.0581e-4 | 266,543 |

Because the TNN can be implemented by a back-to-back connection of an INN and a pre-trained FNN, we can make nine configurations of hybrid TNNs, as shown in Figure 5(b), using MLP, CNN, and LSTM. The network structure and hyperparameters of the pre-trained FNN are the same as those used for the implementation of the FNN mentioned above. To train the TNNs, only network structure and hyperparameters of the INN inside the TNN are optimized. In Sections 3 of Supplementary Information, we present more details about the network structure and hyperparameters of nine TNNs such as the number of hidden layers, the number of neurons in each hidden layer, and a type of the activation function.

Table 2 shows performance comparison of nine network strctures of hybrid TNNs when the number of the thin-film layer is 20. The accuracy of the prediction results of the nine combinations of TNNs is evaluated by calculating R2 scores and MSE between the input transmission spectrum and the predicted transmission spectrum obtained by the inversely-designed layer thickness among 20,000 test datasets. It seems to be uncertainty in generalizing the network's performance based on a single training session. Therefore, we employed the same calculations three times for each of the nine TNN algorithms and summarized the averaged results in Table 2. In Table 2, using LSTM in the pre-trained FNN part rather than in the INN part is better to improve the accuracy of the TNN. When LSTM is applied to an INN, the size of the input data, which corresponds to an $1 \times 401$ vector, will be large for the LSTM algorithm to effectively integrate past and present training datasets. In this sense, the MLP-LSTM-based TNN has a better accuracy than the MLP-MLP-based TNN, but the LSTM-MLP-based TNN has the worse accuracy than the MLP-MLP-based TNN. In contrast, using CNN in the INN part rather than in the pre-trained FNN part is better to improve the accuracy of the TNN. When CNN is applied to an INN, the input data corresponds to nearly continuous optical transmission spectrum that can have very high correlation between adjacent input values like images. Correspondingly, the inherent characteristics of the CNN algorithm, which effectively extract the local correlation between neurons of adjacent layers, resulting in more accurate inverse design results [23,24] and contributing to the accuracy improvement of TNNs. In this regard, the CNN-MLP-based TNN has the better accuracy than the MLP-MLP-based TNN, but the MLP-CNN-based TNN has the worse accuracy than the MLP-MLP-based TNN.

In Table 2, it is found that the LSTM-LSTM-based TNN has the highest accuracy among nine configurations of TNNs. The MSE of the LSTM-LSTM-based TNN is reduced by a factor of four compared to that of the MLP-MLP-based TNN. This superiority of the the LSTM-LSTM-based TNN will result from the inherent memory property of the LSTM algorithm. However, the LSTM-LSTM-based TNN takes the longest training time, where the training time of the LSTM-LSTM-based TNN is fourteen times larger than that of the MLP-MLP-based TNN. The long training time of the LSTM-LSTM-based TNN is primarily attributed to the INN part, where the LSTM must process a long sequence of input transmission spectra represented as a 1 x 401 vector for inverse design. In addition, it is noticeable that the CNN-LSTM-based TNN has the second highest accuracy while its training time increases only by a factor of 3.7 compared to the MLP-MLP-based TNN. Because the CNN-LSTM-based TNN can be considered to integrate the strengths of the CNN-based INN and the LSTM-based FNN, it can provide an optimal solution to the configuration of the TNN in terms of accuracy and speed.

## 4.3. Performance variation of nine hybrid TNNs as a function of the number of multilayer in the thin film

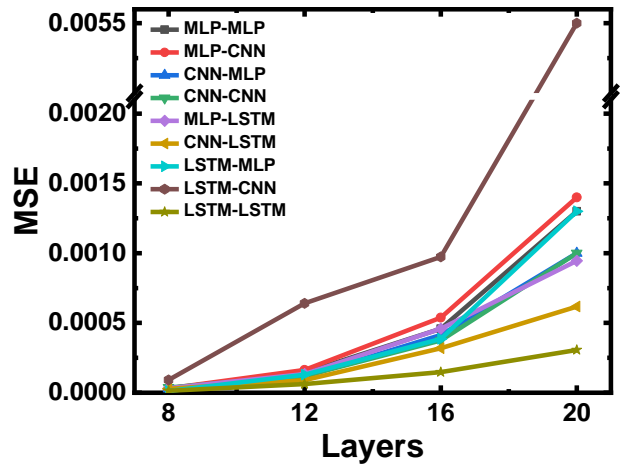

**Fig. 6.** Changes in the MSE of nine network structures of hybrid TNNs as a function of the number of multilayer in the thin film when the size of datasets is fixed as 100,000 for 8-, 12-, 16-, and 20-layer thin films.

Figure 6 shows the changes in the MSE of nine network structures of hybrid TNNs as a function of the number of multilayer in the thin film when the size of datasets is fixed as 100,000. In case of the 8-, 12-, and 16-layer thin films, total 100,000 datasets are calculated based on the TMM depending on the number of thin-film layer. Then, the network structures and hyperparameters of each TNN in the 8-, 12-, and 16-layer thin films are separately optimized in the same manner with the case of the 20-layer thin film. When the thickness range of 30-70 nm with the interval of 1.0 nm is considered, increasing the number of the thin-film layer from 8 to 20 results in dramatic increment of possible layer configurations from $41^8$ to $41^{20}$. Because the number of the training dataset is the same as 60,000 for the 8-, 12-, 16-, and 20-layer thin films, the accuracy of the TNN will be degraded as the number of the thin-film layer increases. In Section 4 of Supplementary Information, we present more details on performance comparison of nine hybrid TNNs at the 8-, 12-, 16-, and 20-layer thin films. In Figure 6, the MSE value obtained by the MLP-MLP-based TNN is already very low for the 8-layer thin film such that the reduction of the MSE obtained by the LSTM-LSTM-based TNN is not so significant. On the contrary, the MSE value obtained by the MLP-MLP-based TNN is relatively high for the 20-layer thin film and the LSTM-LSTM-based TNN provides a significant reduction of the MSE. Hence, our proposed CNN-LSTM-based or LSTM-LSTM-based TNN will be more beneficial when the number of thin-film layers becomes very large.

### 4.4. Performance analysis and interpretation for CNN-LSTM-based TNNs

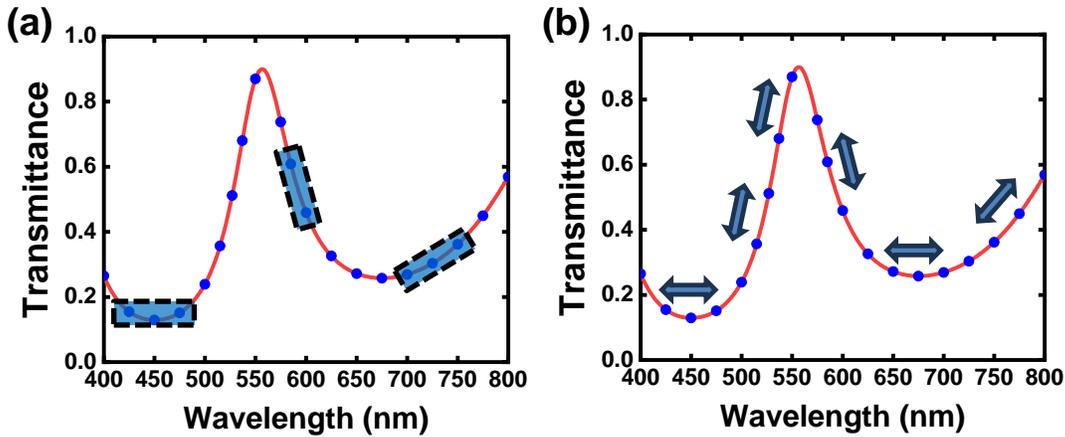

**Fig. 7.** Schematic diagram of spectrum learning processes of (a) CNN and (b) LSTM.

Figure 7 shows schematic diagrams of the spectrum learning process employed by CNN and LSTM algorithms, respectively. As shown in Figure 7(a), CNNs extract features by analyzing input data locally. In the case of transmission spectra, which exhibit continuous variations across different wavelengths, the filters of CNN can efficiently capture these patterns and extract relevant spectral features. This localized feature extraction capability significantly enhances the model's ability to recognize key characteristics within the spectral data. In the case of LSTMs, their networks are specifically designed to manage long-term dependencies through their cell state and gating mechanisms. As indicated in Figure 7(b), this architecture allows LSTMs to learn complex relationships within sequential data, where the transmission at a particular wavelength may be influenced by wavelengths that are spectrally distant. By effectively capturing these inter-wavelength dependencies, LSTM can more accurately learn the variations in spectral response as a function of layer thickness, ultimately improving the predictive performance of the TNN. However, the efficiency of learning from input data diminishes when the data length is excessively long, resulting in prolonged training time.

Figure 8 shows schematic diagrams of the learning processes associated with the CNN and LSTM algorithms in the context of layer thickness. As depicted in Figure 8(a), CNNs are specialized in learning local features and, therefore, cannot effectively capture the overall structural context of the multilayer systems. This limitation arises from the fact that the physical properties of individual layers are not inherently interdependent but rather exist as separate entities, which poses challenges for CNNs in understanding these relationships. Additionally, the layer thickness exhibit sequential relationships, which impose limitations on CNN's capacity to learn them effectively. In contrast, Figure 8(b) demonstrates that LSTMs, which are designed to capture interlayer dependencies, can effectively learn the sequential stacking of layer thicknesses in multilayer thin films. This capability allows LSTMs to make more accurate predictions regarding layer thickness, as variations in the thickness of a specific layer significantly impact the overall shape of the transmission spectrum. Consequently, LSTMs are particularly well-suited for modeling these complex relationships.

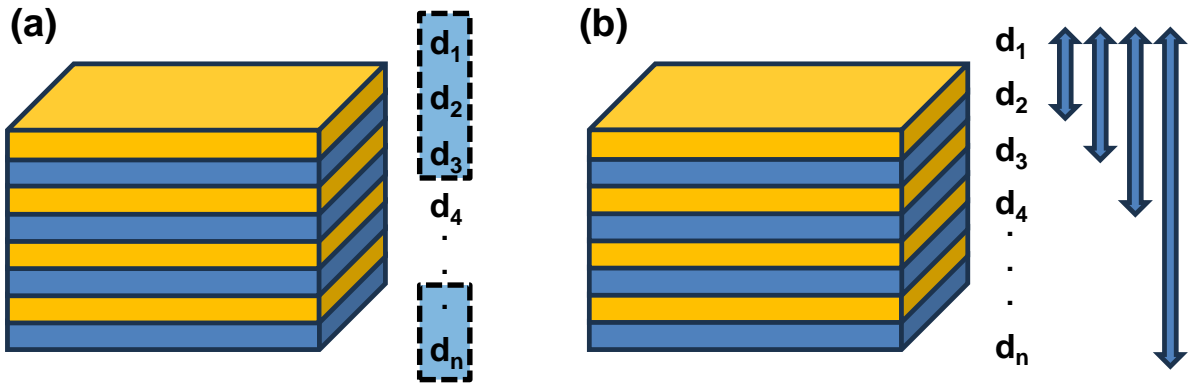

**Fig. 8.** Schematic diagrams of the learning processes associated with the CNN and LSTM algorithms in the context of layer thickness.

## 5. Conclusion

We investigated a deep-learning-based inverse design model for $SiO_2/TiO_2$ multilayer thin films, leveraging the structural parameter of layer thicknesses and the optical properties of the transmission spectrum. The datasets were generated through numerical calculations based on the TMM. In general, the MLP-MLP-based TNN was widely used to perform the inverse design, where the one-to-many mapping problem was avoided. We propose to use a combination of MLP, CNN, and LSTM as the INN and the pre-trained FNN part of TNNs. Nine different network structures of TNNs were implemented and their performance were compared in terms of accuracy and speed. According to experimental results obtained at the 20-layer thin film, the LSTM-LSTM-based TNN had the highest accuracy, which was improved by a factor of four compared to the MLP-MLP-based TNN. On the other hand, the LSTM-LSTM-based TNN had the longest training time, which was longer by a factor of fourteen than the training time of the MLP-MLP-based TNN. In addition, the CNN-LSTM-based TNN had the second highest accuracy while its training time increased only by a factor of 3.7 compared to the MLP-MLP-based TNN. We found that the CNN-LSTM-based TNN would be an optimal solution to the configuration of the TNN in terms of accuracy and speed because it could integrate the strengths of the CNN-based INN and the LSTM-based FNN. The inherent memory characteristics of the LSTM algorithm helped to obtain more accurate forward prediction results by effectively integrating past and present training datasets. The inherent convolution characteristics of the CNN algorithm contributed to improving inverse design results by effectively extracting the local correlation of continuous transmission spectrum. We also found that our proposed CNN-LSTM-based TNN will be more beneficial when the number of thin-film layers becomes very large and conventional MLP-MLP-based TNN cannot provide good accuracy.

## Funding

National Research Foundation of Korea (2021R1F1A1062591).

## CRediT authorship contribution statements

**Uijun Jung:** Conceptualization, Methodology, Software, Visualization, Writing - original draft, Writing – review & editing. **Deokho Jang:** Conceptualization, Methodology, Software, Writing – review & editing. **Sungchul Kim:** Conceptualization, Methodology, Writing - review & editing. **Jungho Kim:** Conceptualization, Methodology, Visualization, Supervision, Writing - original draft, Writing - review & editing.

## Declaration of competing interest

The authors declare that they have no known competing financial interests or personal relationships that could have appeared to influence the work reported in this paper.

## Data availability

Data underlying the results presented in this paper are not publicly available at this time but may be obtained from the authors upon reasonable request.

## Acknowledgements

This research was supported by the Basic Science Research Program (NRF-2021R1F1A1062591) funded by the National Research Foundation of Korea.

## Supplemental document

See Supplement for supporting content.

# Improving the performance of optical inverse design of multilayer thin films using CNN-LSTM tandem neural networks: Supplement

Uijun Jung[1], Deokho Jang[1], Sungchul Kim[2], and Jungho Kim[1, *]

[1]Department of Information Display, Kyung Hee University, Seoul, 02447, Republic of Korea
[2]Department of Information and Communications Engineering, Myongji University, Yongin-si, 17058, Republic of Korea

*junghokim@khu.ac.kr

**1. Transfer matrix method (TMM) and refractive index spectra of the materials**

*A. Theoretical model of the TMM*

The light is assumed to be normally incident into the multilayer thin film. When $d_j$ represents the thickness of the $j$-th layer, the wave vector can be written as $k_{x,j} = k_{x,0}$ because the transverse wave vector is continuous at the interface [1]. Then, the wave vectors can be written as

$$K_j = \left( K_{x,j},\ 0,\ K_{z,j} \right),$$
$$K_{x,j} = K_{x,0} = n_0 \sin(\theta) k_0,$$
$$K_{z,j} = n_j \cos(\theta_j) k_0 = \left( p_j + i q_j \right) k_0 \quad (1)$$

It can be written as $p_j + i q_j = \sqrt{(n_j + i k_j)^2 - (n_0 \sin\theta)^2}$ by dispersion relation of $k_{x,j}^2 + k_{z,j}^2 = \tilde{n}_j^2 k_0^2$. The polarization-dependent reflection/transmission coefficient can be determined by the Fresnel equation, which is given by

$$r_{jk}^s = \frac{\tilde{n}_j \cos(\theta_j) - \tilde{n}_k \cos(\theta_k)}{\tilde{n}_j \cos(\theta_j) + \tilde{n}_k \cos(\theta_k)},\ t_{jk}^s = \frac{2\tilde{n}_j \cos(\theta_j)}{\tilde{n}_j \cos(\theta_j) + \tilde{n}_k \cos(\theta_k)} \quad (2)$$

The electric field magnitude at the interface between neighboring layers is calculated through interface matrix $I$, and the electric field magnitude inside the layer is calculated through the layer matrix $L$ as follows [1].

$$\begin{bmatrix} E_{JR}^+ \\ E_{JR}^- \end{bmatrix} = I^{jk} \begin{bmatrix} E_{KL}^+ \\ E_{KL}^- \end{bmatrix} = \frac{1}{t_{jk}} \begin{bmatrix} 1 & r_{jk} \\ r_{jk} & 1 \end{bmatrix} \begin{bmatrix} E_{KL}^+ \\ E_{KL}^- \end{bmatrix} \quad (3)$$

$$\begin{bmatrix} E_{JR}^+ \\ E_{JR}^- \end{bmatrix} = \mathbf{L}(\mathbf{d}_j) \begin{bmatrix} E_{jL}^+ \\ E_{jL}^- \end{bmatrix} = \begin{bmatrix} e^{-ik_{z,j}d_j} & 0 \\ 0 & e^{ik_{z,j}d_j} \end{bmatrix} \begin{bmatrix} E_{jL}^+ \\ E_{jL}^- \end{bmatrix} \quad (4)$$

Then, the amplitude scattering matrix $S$ that relates the electric field amplitude between the input ambient ($j=0$) and the $j$-th layer is given by

$$S^{0/j} = \begin{bmatrix} S_{11}^{0/j} & S_{12}^{0/j} \\ S_{21}^{0/j} & S_{22}^{0/j} \end{bmatrix} = I^{01} L^1 I^{12} \cdots L^{(j-1)} I^{(j-1)j} \quad (5)$$

Finally, the total transmission/reflection coefficient for the electric field amplitudes is expressed as follows [1].

$$r_{0(M+1)} = \frac{E_{0R}^{-}}{E_{0R}^{+}} = \frac{S_{21}^{0/(M+1)}}{S_{11}^{0/(M+1)}} \quad (6)$$

$$t_{0(M+1)} = \frac{E_{(M+1)L}^{+}}{E_{0R}^{+}} = \frac{1}{S_{11}^{0/(M+1)}} \quad (7)$$

*B. Refractive index spectra of SiO$_2$ and TiO$_2$*

The refractive index spectra of SiO$_2$ and TiO$_2$ are taken from the literature [2]. Because the extinction coefficients of SiO$_2$ and TiO$_2$ are negligible in the wavelength range between 400 and 800 nm, we only consider the refractive index of SiO$_2$ and TiO$_2$ in the calculation of transmission spectrum.

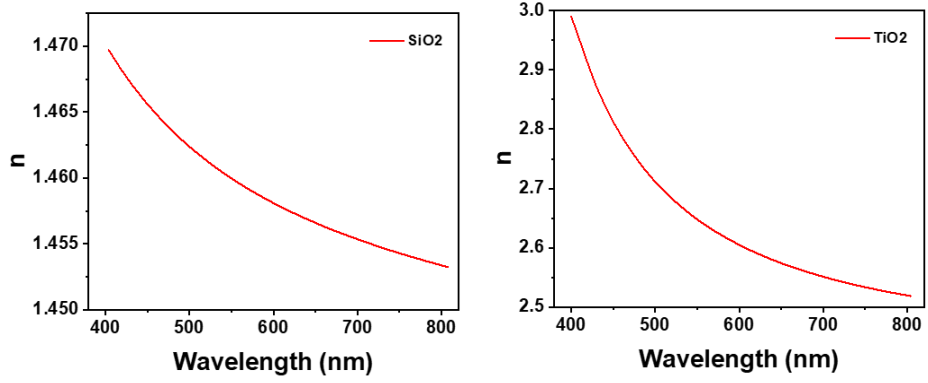

Fig. S1. Refractive index spectra of SiO$_2$ and TiO$_2$ used in the calculation.

**2. Details about three forward neural networks (FNNs)**

*A. Network structures and hyperparameters*

Table S1. Network structures of three FNNs (20-layers).

| Algorithm | FNN |
| --- | --- |
| MLP | 100-200-300-400-401 (Dense layer) |
| CNN | 10-20-40(Conv1D layer)- 100-200-300-400-401 (Fully connected layer) |
| LSTM | 20-100-200-401 (LSTM layer) |

Table S2. Hyperparameters common to the MLP-, CNN-, and LSTM-based FNNs (20-layers).

| Hyperparameter | Setting |
| --- | --- |
| Epochs | 500 |
| Batch size | 16 |
| Activation function | Leaky Relu, Sigmoid (output layer) |
| Optimizer | Adam |
| Loss function | Mean squared error |
| Learning rate | 1e-4 |

Table S3. Hyperparameters that only apply to the CNN-based FNN (20-layers).

| Hyperparameter | Setting |
| --- | --- |
| Kernel size | 3 |
| Padding | Same |
| Activation | Relu |
| Pooling | Max pooling (size=2) |

## B. Learning loss curve and test example

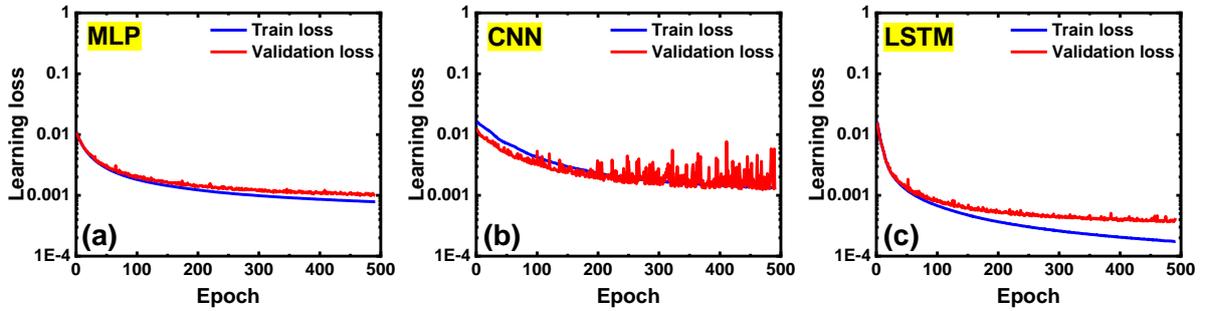

Fig. S2. Comparison of the learning loss curves between three FNNs. (a) MLP-based FNN, (b) CNN-based FNN, (c) LSTM-based FNN. Train loss (blue) and validation loss (red) are designated.

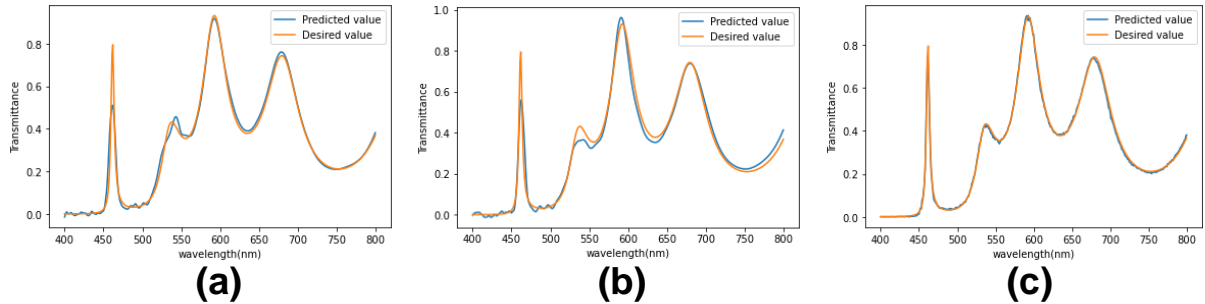

Fig. S3. Test example of three FNN. (a) MLP-based FNN, (b) CNN-based FNN, (c) LSTM-based FNN. Target (orange) and predicted spectrum (blue) are represented.

## 3. Details about nine network structures of tandem neural networks (TNNs)

*A. Network structures and hyperparameters*

Table S4. Network structures of TNNs (20-layers).

| Algorithm | INN part | Pre-trained FNN part |
|---|---|---|
| MLP | 800-400-200-100-20 (Dense layer) | 100-200-300-400-401 (Dense layer) |
| CNN | 30-60-120(Conv1D layer)- 2000-1000-500-100-20 (Fully connected layer) | 10-20-40(Conv1D layer)- 100-200-300-400-401 (Fully connected layer) |
| LSTM | 100-50-30-20 (LSTM layer) | 20-100-200-401 (LSTM layer) |

Table S5. Hyperparameters that only apply to the CNN used in the TNN (20-layers).

|  | INN part | Pre-trained FNN part |
|---|---|---|
| Kernel size | 11 | 3 |
| Padding | Same | Same |
| Activation | Relu | Relu |
| Pooling | Max pooling (size=2) | Max pooling (size=2) |
| Batch normalization layer is used after every Conv1D layer in the CNN. | | |

Table S6. Hyperparameters common to the MLP-, CNN-, and LSTM-based TNNs (20-layers).

| Hyperparameter | Setting |
|---|---|
| Epochs | 1000 |
| Batch size | 16 |
| Activation function | Leaky Relu, Sigmoid (output layer) |
| Optimizer | Adam |
| Loss function | Mean squared error |
| Learning rate | 1e-4 |
| Patience | 200 |

*B. Learning loss curve and test example*

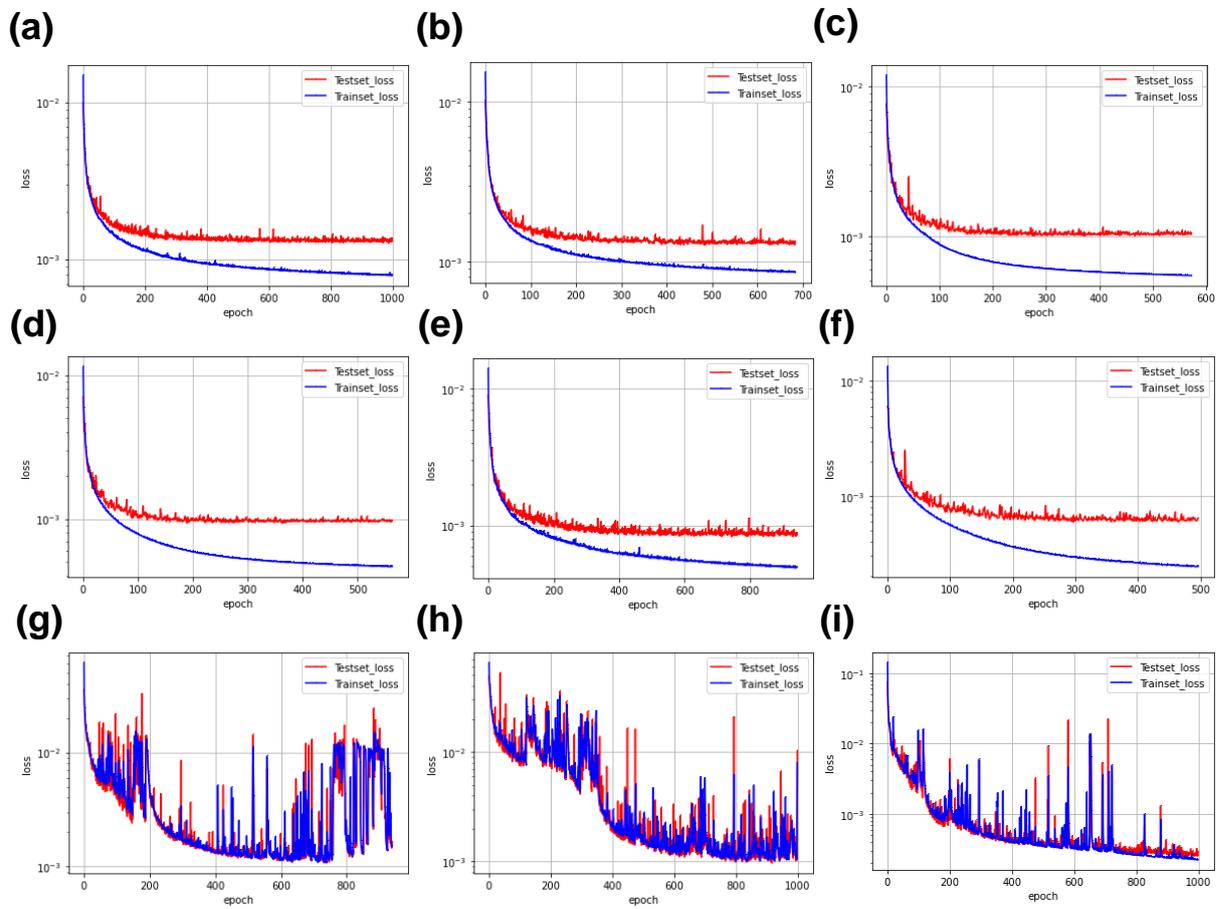

Fig. S4. Comparison of the learning loss curves between nine network structures of TNNs (20-layers). (a) MLP-MLP, (b) MLP-CNN, (c) CNN-MLP, (d) CNN-CNN, (e) MLP-LSTM, (f) CNN-LSTM, (g) LSTM-MLP, (h) LSTM-CNN, (i) LSTM-LSTM based TNNs.

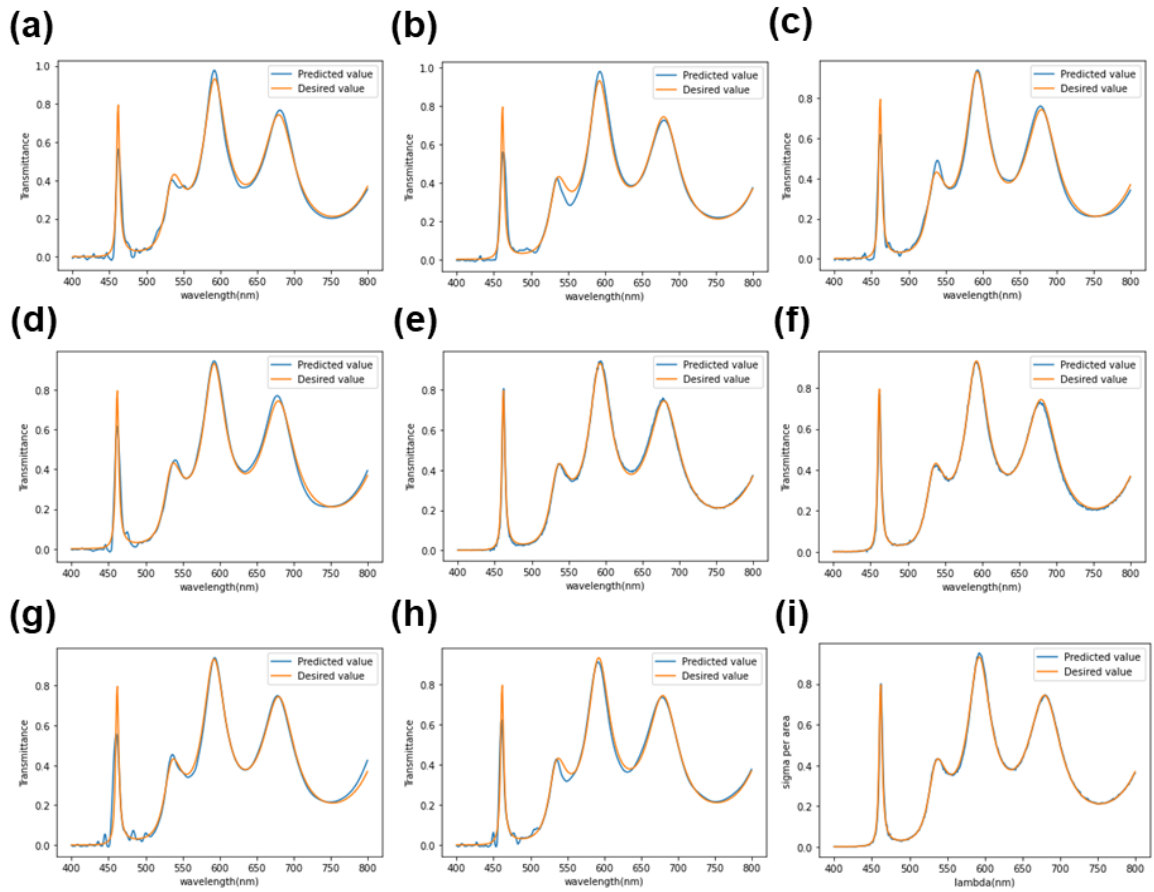

Fig. S5. Test example of nine network structures of TNNs (20-layers). (a) MLP-MLP, (b) MLP-CNN, (c) CNN-MLP, (d) CNN-CNN, (e) MLP-LSTM, (f) CNN-LSTM, (g) LSTM-MLP, (h) LSTM-CNN, (i) LSTM-LSTM based TNNs.

**4. Details on performance comparison of nine hybrid TNNs at the 8-, 12-, 16-, and 20-layer thin films. (M, C, and L, denoting MLP, CNN, and LSTM, respectively)**

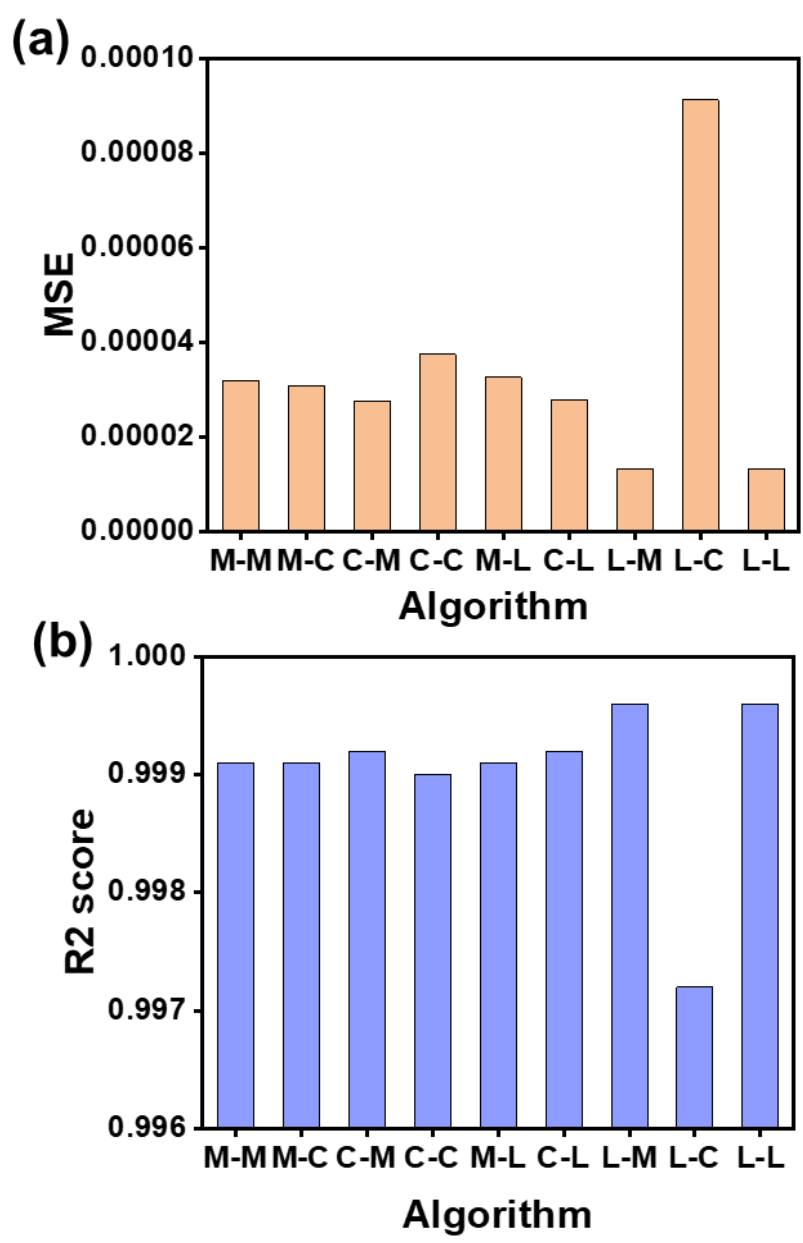

Fig. S6. Comparison of (a) MSE values and (b) R2 scores of nine hybrid TNNs at the 8-layer thin film.

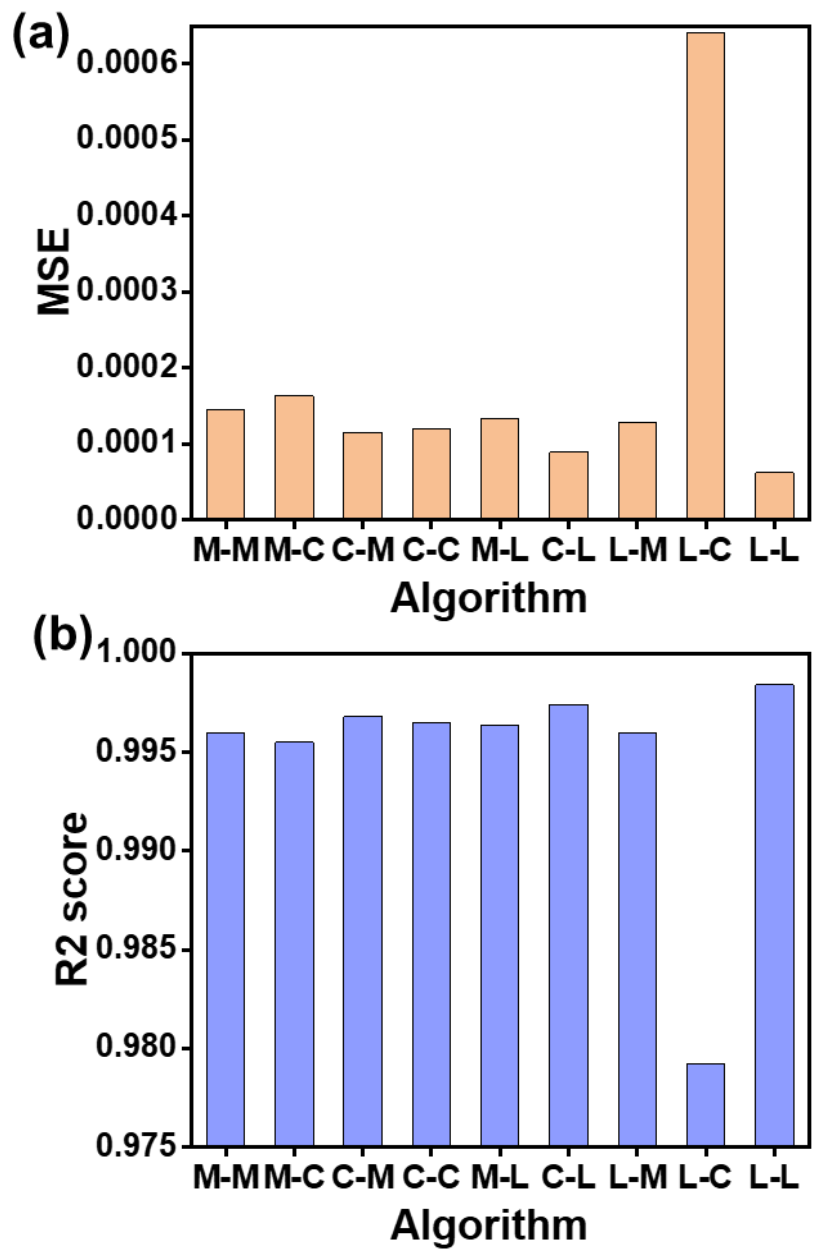

Fig. S7. Comparison of (a) MSE values and (b) R2 scores of nine hybrid TNNs at the 12-layer thin film.

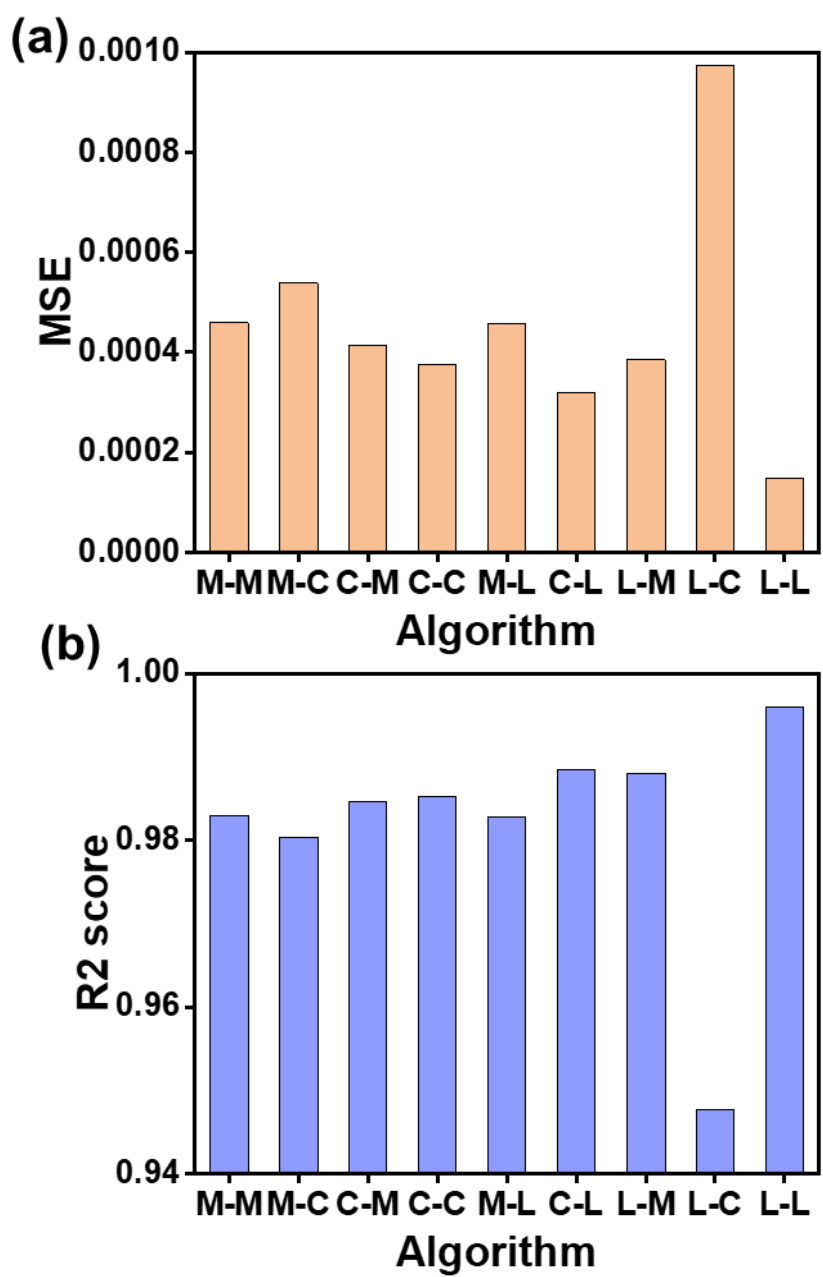

Fig. S8. Comparison of (a) MSE values and (b) R2 scores of nine hybrid TNNs at the 16-layer thin film.

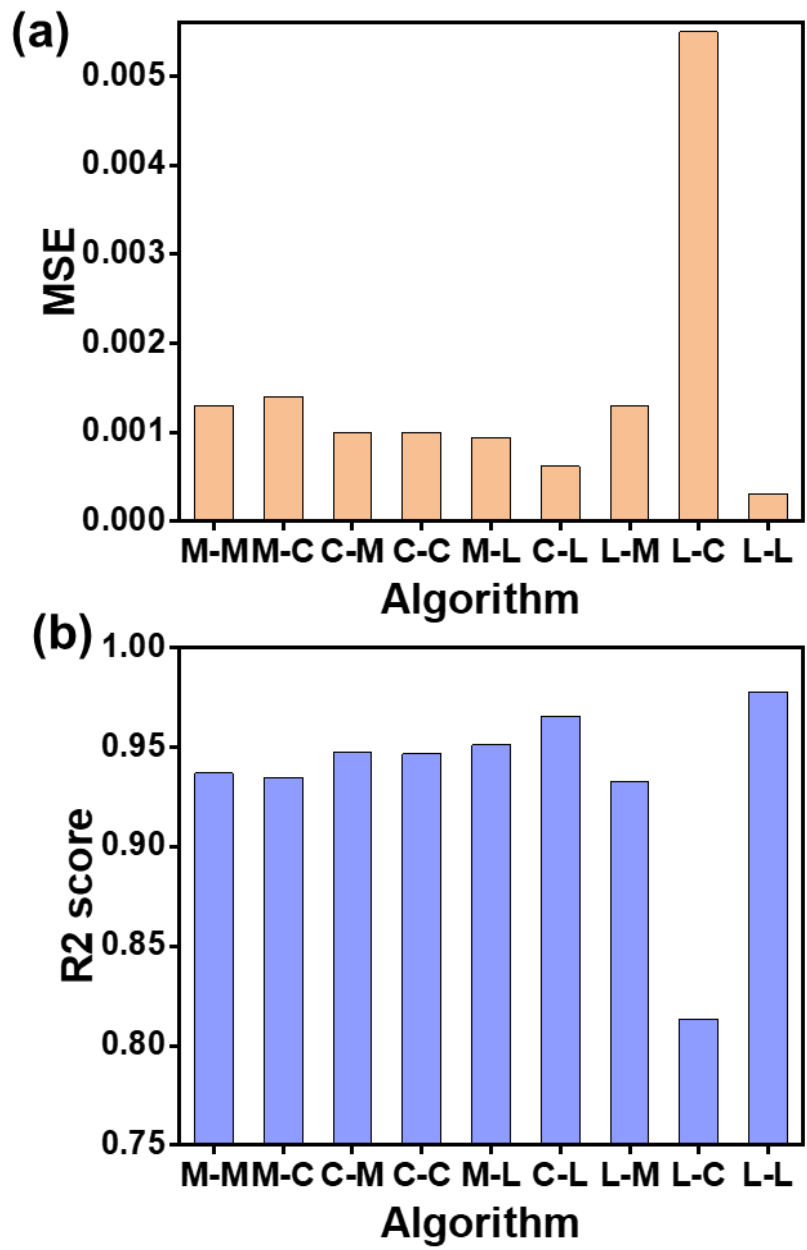

Fig. S9. Comparison of (a) MSE values and (b) R2 scores of nine hybrid TNNs at the 20-layer thin film.

## 5. Comparison of optimization time between TMM-based traditional genetic algorithm (GA) and FNN-assisted GA

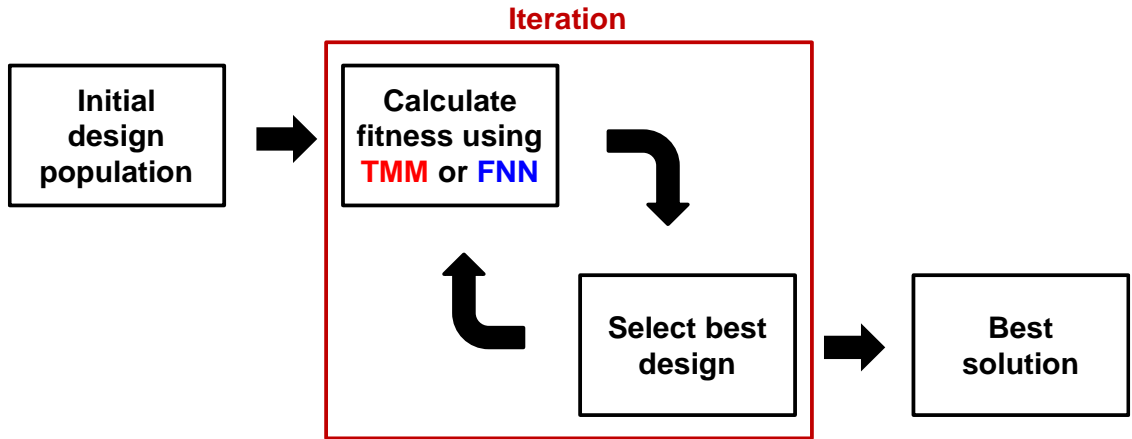

Fig. S10. Schematic diagram of the two GAs for inverse design.

Table S7. Hyperparameters common to the TMM-, FNN- based GA (20-layers).

| Hyperparameter | Setting |
| --- | --- |
| Population size | 200 |
| Number of generations | 500 |
| Mutation rate | 0.1 |
| Selected percentage | 0.1 |
| Fitness score | Mean squared error |

We compared the inverse design results based on a representative traditional algorithm, the GA. As shown in Figure S10, one approach utilized a TMM simulator to compare the fitness score (MSE) between the target spectrum and the predicted spectrum, while the other replaced the TMM simulator with a FNN and compared the fitness score. In Table S7, the hyperparameters were set identically in both cases. Figure S11 shows MSE values and optimization time of the traditional and FNN-assisted GA. In the optimization based on the FNN-assisted GA, an MLP-based FNN used for training the TNN was employed. As mentioned in the manuscript, the generation time for the 100,000 datasets, consisting of 20-layers thin-film thicknesses and their transmission spectra, was 6 hours and 20 minutes. The training time for the MLP-based FNN was 1 hour and 45 minutes. The optimization results showed that the FNN-assisted GA took 45 minutes, whereas the traditional GA took 54 minutes, indicating that the FNN-assisted GA achieved faster training.

In this study, we trained the model using the transmittance and thickness of multilayer films composed of $SiO_2$ and $TiO_2$. The TMM was used to compute the transmittance and reflectance of the multilayer films. However, since TMM does not involve highly complex computational processes, using traditional algorithms for optimization results in a relatively small computational time. In contrast, if traditional algorithms incorporate methods that require more complex calculations, such as rigorous coupled-wave analysis (RCWA) or finite-difference time-domain (FDTD), the optimization time can become significantly longer. In such cases, employing neural networks for training could enable faster inverse design.

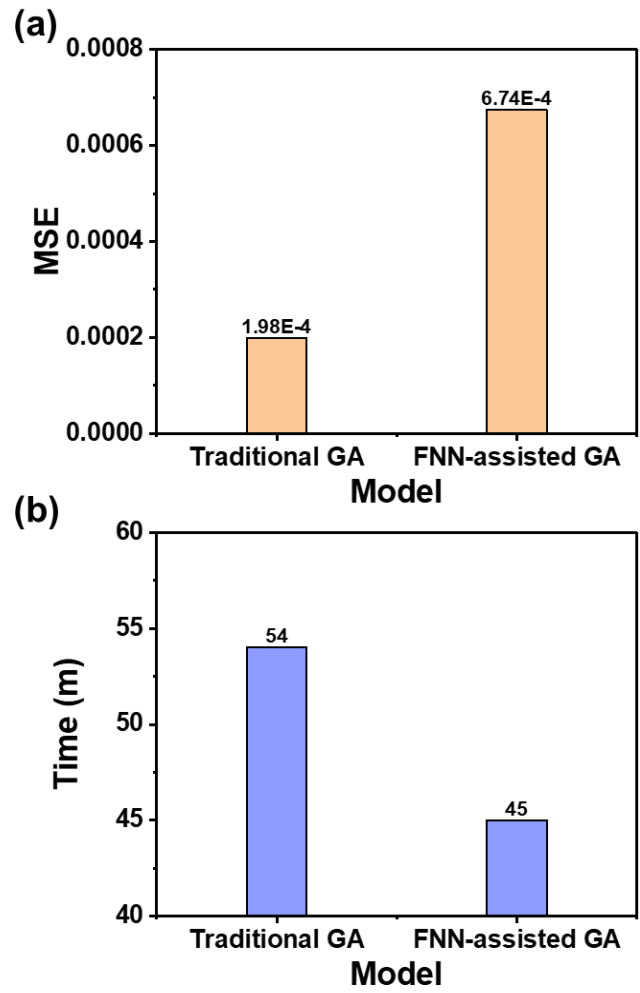

Fig. S11. Comparison of (a) MSE values and (b) optimization times between the TMM-based and FNN-assisted GAs to perform the inverse design of a 20-layer thin film for a given target transmission spectrum.